# A Novel Pearson Correlation-Based Merging Algorithm for Robust Distributed Machine Learning with Heterogeneous Data

Mohammad Ghabel Rahmat
Department of Computer
Engineering, Karaj Branch
Islamic Azad University
Karaj, Iran
GhabelRahmat.m@gmail.com

Majid Khalilian
Department of Computer
Engineering, Karaj Branch
Islamic Azad University
Karaj, Iran
khalilian@kiau.ac.ir

*Abstract*— Federated learning faces significant challenges in scenarios with heterogeneous data distributions and adverse network conditions, such as delays, packet loss, and data poisoning attacks. This paper proposes a novel method based on the SCAFFOLD algorithm to improve the quality of local updates and enhance the robustness of the global model. The key idea is to form intermediary nodes by merging local models with high similarity, using the Pearson correlation coefficient as a similarity measure. The proposed merging algorithm reduces the number of local nodes while maintaining the accuracy of the global model, effectively addressing communication overhead and bandwidth consumption. Experimental results on the MNIST dataset under simulated federated learning scenarios demonstrate the method's effectiveness. After 10 rounds of training using a CNN model, the proposed approach achieved accuracies of 0.82, 0.73, and 0.66 under normal conditions, packet loss and data poisoning attacks, respectively, outperforming the baseline SCAFFOLD algorithm. These results highlight the potential of the proposed method to improve efficiency and resilience in federated learning systems.

I. INTRODUCTION

As information technology, the Internet, social networks, and Internet of Things (IoT) devices advance, vast amounts of data are generated. Businesses and organizations must analyze and process this data, necessitating increased computing power and specialized expertise. The primary challenges in processing data or applying machine learning are encapsulated in the "5Vs" of big data.[1].

Distributed machine learning effectively addresses significant challenges by employing multiple systems in parallel, rather than relying on a single central system to process all data and perform computations. This approach offers several advantages, including reduced processing time, enhanced scalability, improved reliability, and cost reduction. For instance, instead of deploying a costly high-performance server, multiple more affordable servers can be utilized.

Distributed machine learning methods are generally classified into five categories: data parallelism, model parallelism, hybrid parallelism, decentralized machine learning, and federated learning. Data parallelism involves distributing data across multiple nodes, with each node training a model replica on its subset of data. Model parallelism partitions the model itself across different nodes, where each node is responsible for computing a portion of the model. Hybrid parallelism combines data and model parallelism to leverage the benefits of both approaches. Decentralized machine learning eliminates the central coordinator, allowing nodes to communicate directly and collaboratively update the model. Federated learning trains models across decentralized devices holding local data samples without exchanging them, thereby preserving data privacy. These categories encompass the primary strategies employed in distributed machine learning. [2]

Considering factors such as data volume, model complexity, hardware limitations, privacy requirements, and communication costs, selecting an appropriate distributed machine learning method is crucial. This article focuses on federated learning, an approach where, instead of transferring data to a central server, models are dispatched to devices housing the data. These models are trained locally on each device, enhancing privacy and reducing data transmission costs. [3-5]

Given that federated learning prioritizes data owners' privacy, it proves highly effective in sectors like healthcare, the Internet of Things (IoT), and finance. In healthcare, federated learning enables hospitals and medical centers to collaboratively train models without sharing sensitive patient data. For instance, tasks such as analyzing electronic health records (EHRs), interpreting medical images (e.g., MRIs), and making clinical predictions (e.g., disease diagnosis) can achieve high accuracy using federated learning. [6-7]

In the realm of the Internet of Things (IoT), federated learning can be applied to smart devices and smart cities. For instance, smart transportation systems and smart energy grids can analyze local data and develop optimal models without transmitting raw data to a central server. This approach enhances data privacy and reduces communication costs.[8]

In the financial sector, federated learning (FL) is instrumental in detecting financial fraud and mitigating credit risks. By enabling collaborative model training across multiple institutions without sharing sensitive customer data, FL allows banks and financial entities to identify fraudulent activities while upholding data privacy. For instance, FL facilitates the detection of complex money laundering patterns that span multiple financial institutions, enhancing the effectiveness of anti-money laundering efforts. Additionally, FL-based models have demonstrated improved performance in credit risk assessment by leveraging diverse data sources without compromising individual privacy.[9]

Federated learning (FL) is designed to protect data owners' privacy, benefiting all participants. Instead of aggregating data in a single location, FL distributes the learning process across multiple devices. Various federated learning techniques have been developed to facilitate this distribution. Considering factors such as data heterogeneity and convergence speed, different strategies for federated learning have been proposed. Data heterogeneity, or non-IID data, poses challenges in federated learning, affecting both convergence speed and model accuracy. To address this, methods like incorporating proximal terms in local optimization, modifying server-side model aggregation, and employing clustered federated learning approaches have been developed. Additionally, techniques such as knowledge distillation have been integrated into federated learning algorithms to handle data heterogeneity, aiming to improve convergence speed and prediction accuracy. [10]

## II. PROBLEM STATEMENT

The Federated Averaging (FedAvg) algorithm is a fundamental strategy in federated learning, designed to address challenges such as minimizing communication overhead and preserving data privacy. In FedAvg, each client device trains a local model on its own data and periodically sends the updated model parameters to a central server. The server aggregates these parameters, typically by computing their average, to update the global model. This process reduces the frequency and volume of data transmission, as only model updates—not raw data—are shared, thereby enhancing privacy and efficiency in distributed machine learning environments. [11]

The Federated Averaging (FedAvg) algorithm faces challenges when dealing with heterogeneous or unbalanced data, as well as limited and unreliable network connections. Data distributed across different devices often exhibit non-independent and non-identically distributed (non-IID) characteristics, leading to diverse data patterns. This variability can diminish the efficiency of the federated model, as devices may adapt to distinct data patterns due to content differences. Additionally, the number of data instances per device can vary significantly; for example, one device might have hundreds of instances, while another has only a few. Such imbalance can cause the model to prioritize data from certain devices over others, thereby reducing overall performance.[12]

Another challenge with the Federated Averaging (FedAvg) algorithm is its slow convergence, particularly in the presence of heterogeneous data. The variability in data distributions across devices can lead to significant differences in local updates, causing instability and slower convergence rates. Additionally, the quality of local updates plays a crucial role in the overall performance of the central model. Devices with computational limitations or insufficient data may produce suboptimal local models. When these inferior updates are aggregated, they can degrade the quality of the global model, especially if a substantial number of devices contribute low-quality updates. Addressing these issues requires strategies that mitigate the effects of data heterogeneity and ensure the reliability of local updates to maintain the integrity and efficiency of the federated learning process.[13]

To address the challenge of data heterogeneity in federated learning, the Federated Proximal (FedProx) algorithm has been introduced. The core concept of FedProx is to maintain local models closer to the central model by incorporating a proximal term into the local objective functions. In the standard Federated Averaging (FedAvg) algorithm, each device updates its local model independently, and these updates are then aggregated by the central server. However, in FedProx, a proximal term is added to the local cost function, which penalizes deviations from the central model. This constraint ensures that local models do not diverge significantly from the global model, even in the presence of heterogeneous or unbalanced data distributions. As a result, devices with diverse data contribute more effectively to the convergence of the central model, enhancing overall performance and stability in federated learning environments.[14]

While the Federated Proximal(FedProx) algorithm effectively addresses data heterogeneity by incorporating a proximal term to keep local models closer to the global model, it may still encounter challenges related to low-quality local updates. Devices with computational limitations or poor-quality data can produce suboptimal local models, which, when aggregated, may adversely affect the global model's performance. However, by appropriately selecting the proximal parameter (μ), the impact of these low-quality updates can be mitigated. A well-chosen μ value constrains the extent to which local updates deviate from the global model, thereby reducing the influence of less reliable local updates on the overall model's convergence and accuracy.

In the Federated Proximal (FedProx) algorithm, the parameter μ, known as the proximal or penalty parameter, regulates the influence of the proximal term added to the local objective functions. This term constrains the divergence between local updates and the global model, ensuring that local training remains aligned with the central model. By appropriately tuning μ, FedProx effectively manages the balance between local optimization and global convergence, thereby enhancing stability and performance in federated learning environments with heterogeneous data.[15]

The scaffold algorithm [26] is proposed to address the client drift problem. It achieves this through two control parameters: local and global. The local parameter ensures that local nodes obtain a better estimate of the true gradient, while the global control parameter coordinates the overall settings of the algorithm to prevent unnecessary deviations. However, when the network is subjected to adverse conditions, such as packet loss and poisoning attacks, the final accuracy decreases because poisoned local nodes send deviating parameters to the central node.

One solution to this issue is to improve the quality of local nodes' data before entering the federated learning rounds by employing methods with high computational complexity, which will be discussed in the next section. However, using these methods imposes significant computational costs on the network and necessitates more advanced solutions for resource management.

Providing a defense mechanism that can mitigate the impact of poisoned nodes and prevent the deviation of the central model from the main learning path without requiring complex processing operations or improving the quality of local node data can be highly effective. In the next section, we will review the methods that have addressed the problem of low data quality in local nodes. Subsequently, the proposed method of this article, which is based on the approach of creating a defense mechanism, is presented. In the following section, we conduct experiments to compare the accuracy of the proposed method with the standard scaffold algorithm under adverse network conditions and data heterogeneity. Finally, we present the conclusion and discuss future work.

### III. BACKGROUND OF STUDY

In relation to the problem of low-quality local updating in federated learning models, there are different approaches, and a brief description of each approach is given below.

In [16], a robust approach for federated learning based on data filtering is presented. In this method, the quality of the central model is improved by identifying and removing noisy and invalid data from local updates. This method identifies noisy data and prevents their aggregation into the central model.

In [17], the main focus is on identifying malicious updates. For this purpose, an algorithm is presented to resist destructive or incompatible updates, which are identified and removed through filtering and matching models. This method uses alignment and distance comparison of models to avoid malicious updates. The disadvantage of this method is that it requires coordination and fine-tuning between devices, which may lead to a decrease in accuracy in unstable networks or when the data is highly heterogeneous.

In [18], the main idea is based on local-level data augmentation. Basically, by using data augmentation at the local level, it helps devices with little data to generate more synthetic data and improve model performance. This is done by generating random and diverse data on local devices. The disadvantage of this method is that data augmentation may produce unrealistic and inconsistent data, which may decrease the quality of the central model instead of improving accuracy. Also, this method may not be suitable for complex and multidimensional data.

In [19], using meta-learning, models are trained to adapt to local data in a personalized manner. In this way, pre-trained models can be fine-tuned to local-specific data. The disadvantage of this method is that it requires higher processing resources and memory to fine-tune the models. Additionally, in cases where the local data is very limited and heterogeneous, this method may not perform well.

In [20], an advanced algorithm called SCAFFOLD is introduced, which aims to improve federated learning performance in heterogeneous data using personalized layers and variance control. The weakness of this method is that it has very high computational complexity, and it also requires significant coordination between devices. Furthermore, the presence of highly heterogeneous data may lead to a decrease in the convergence speed of the model.

A mechanism for intelligent weighting of local updates based on data quality is presented in [21]. Updates with higher quality are given more weight to improve the accuracy of the central model. The disadvantage of this method is that determining the appropriate criteria for evaluating data quality and adjusting the weights adds more complexity to the federated learning process and may also lead to unfavorable results in the presence of unbalanced data.

In [22], the problem of federated learning being faced with noisy and inaccurate data is investigated. Due to the nature of federated learning, local data on different devices may vary in quality, and updates originating from noisy data can negatively affect the central model. A noise-tolerant method for aggregating updates is presented, which enables the central model to achieve high accuracy even in the presence of noisy and inaccurate data. This method ensures that the central model maintains its accuracy even when local data is noisy or of low quality. In this way, the system becomes resistant to malicious data or devices generating inaccurate and unreliable data. Additionally, this method does not require significant changes to local devices and is mainly implemented at the aggregation level of the central model. The main disadvantage of this method is that it requires precise criteria for noise detection.

In [23], a mechanism for monitoring the quality of updates is presented, providing feedback to local devices to improve their processes and send better updates when issues are identified. The update quality monitoring and feedback system is a practical tool for improving the accuracy and stability of federated learning, as it prevents invalid updates from entering the central model.

The disadvantage of this method is that adding the monitoring and feedback layer may increase computational costs and cause problems in situations where network bandwidth is limited. Furthermore, this mechanism requires detailed configurations, which may be complex to implement.

The main challenge of these approaches lies in finding a balance between the quality of updates, the consumption of computing resources, and maintaining security and privacy. Many of these methods perform well in simulated conditions, but their practical application in large and complex networks requires more careful investigation. Furthermore, the heterogeneity of data and network conditions remains a significant obstacle. Developing methods that can work equally effectively in both limited-bandwidth network conditions and heterogeneous data environments continues to be a major challenge.

Federated learning requires coordination among a large number of devices, each of which has different data and may operate under varying network conditions such as limited bandwidth or high latency. These characteristics make some approaches unable to achieve the same quality and accuracy observed under simulation conditions in the real world.

If the unadjusted federated model uses the updates from each node equally, the result can be a hybrid model that does not accurately represent any of these groups and whose predictions lack the desired accuracy. The adaptive solutions proposed in the reviewed methods aim to provide each node with a version of the model that aligns with its local data by identifying inconsistent data or personalizing the models.

The challenge of this approach is that implementing model personalization requires additional computing resources and can be inefficient for resource-constrained networks. Furthermore, if the personalization settings are too detailed and unique for each node, optimizing the overall federated model becomes significantly more difficult.

### IV. PROPOSED METHOD

In this paper, we present a method based on the Scaffold algorithm to address the problem of low-quality updates in local federated learning nodes, ensuring robustness under conditions of data heterogeneity, network delay, and data poisoning attacks. To this end, we first provide brief explanations of each concept related to the proposed method.

*A. Non-IID data*

In federated learning, the concept of Non-IID data (non-independent and non-identically distributed) is defined because the data is generated from multiple sources under different conditions. For example, IoT sensors in various devices may each produce a specific distribution of data. In [24], Non-IID data refers to data whose distribution differs among clients (local nodes). Such data closely resemble real-world conditions due to variations in user behavior, device characteristics, or environmental conditions. For instance, different users may have distinct interests, writing styles, or environmental factors, resulting in statistically diverse data.

Given that the data used in federated learning in the real world is inherently non-independent and non-uniformly distributed, research on federated learning algorithms often involves creating datasets with induced heterogeneity. This is achieved through operations such as unbalanced allocation of data classes to local nodes, assigning different amounts of data to local nodes, adding noise by modifying the data, and selecting a subset of features to assign to local nodes.

*B. Scaffold Algorithm*

In the Scaffold algorithm presented in [25], an attempt has been made to solve client drift problems, improve the convergence rate, reduce the variance in gradients, and leverage the similarity of the data across local nodes. Client drift refers to the difference between the local gradients of the nodes and the desired gradient of the central model. This phenomenon occurs when the data of local nodes exhibit heterogeneity, meaning the distribution of data among different local nodes is not uniform. As a result, each node independently optimizes its objective function, which typically does not align with the optimal path of the central model. This mismatch causes the updates to accumulate heterogeneously in the central model, leading to slow convergence or failure to achieve the desired results in the federated learning algorithm.

In the Scaffold algorithm, a variance control mechanism is employed to reduce client drift. This mechanism includes defining a parameter called the Local Control Variate for each node and a Central Variance Control (Global Control Variate) on the server. These parameters measure the difference between the local gradient and the central gradient and adjust their effects on the gradients of the local nodes.

*C. Data poisoning attacks*

Data Poisoning Attacks are a type of attack in machine learning where an attacker attempts to degrade the performance of a model by injecting manipulated data into the model's training process or altering it to cause intentional errors in predictions. These attacks are particularly significant in federated learning and distributed learning scenarios, where data originates from multiple and potentially unreliable sources.

Some types of these attacks include feature manipulation, label manipulation, label flipping, and the injection of fake data [26].

In federated learning, data poisoning can occur at local nodes (clients) because nodes have access to local data, which is not directly controlled by the central server. By modifying local data, an attacker can manipulate the model parameters, causing the final model to deviate from its intended objectives. Due to the distributed nature of federated learning and the central server's lack of access to raw data, detecting such attacks becomes more challenging.

In general, poisoning attacks in federated learning are categorized into two types: Data Poisoning and Model Poisoning. Each employs different methods to manipulate or disrupt the model.

In data poisoning attacks, the attacker manipulates the local training data used for the model. This manipulation may involve altering features, flipping labels, or adding fake data.

In model poisoning, the attacker directly intervenes in the model parameter updates instead of manipulating the data. This type of attack typically occurs during the model update phase, after gradients are calculated. The attacker deliberately modifies the gradients to compromise the final model [27].

Approaches to counter data poisoning attacks include detecting and removing malicious data or employing techniques such as normalizing gradients or weighting nodes based on their accuracy and alignment with the central model.

*D. Proposed Algorithm*

The main idea of the proposed Algorithm is based on the steps of the SCAFFOLD algorithm, with the key difference being the introduction of a merging technique during the training phase of local nodes. After a certain number of federated learning rounds, nodes with similar model weight and bias parameters are identified using the Pearson correlation coefficient and a similarity matrix. These nodes are grouped into pairs and merged to form intermediary nodes. Subsequently, the algorithm involves sending parameters from the central model to the local nodes, training the local nodes, and mitigating client drift based on the parameters of the intermediary nodes.

The merging of similar nodes reduces the number of active nodes, leading to lower network communication overhead, which is especially beneficial in environments with limited bandwidth or high latency. By aggregating similar data, fluctuations caused by heterogeneous datasets are minimized, improving model convergence rates and overall accuracy. Additionally, merging nodes decreases the number of local models requiring training, thereby reducing computational resource consumption and enhancing system productivity.

Nodes that do not meet the similarity threshold remain independent, allowing for better management of unique nodes without adversely affecting the overall model. Furthermore, merging similar nodes can dilute the impact of malicious nodes, as the influence of poisoned data from malicious nodes is lessened when combined with benign nodes, thus safeguarding the central model's integrity.

Adding intermediary nodes between the central model and local nodes enhances scalability, strengthens resilience against data poisoning attacks, and ensures robustness against various network issues, including packet loss, network delays, and data inhomogeneity. This method, while retaining the advantages of the SCAFFOLD algorithm in handling data heterogeneity, reduces computational costs by decreasing the frequency of updates sent to the server. Additionally, it can improve the central model's accuracy.

In summary, the proposed algorithm leverages the scaffold algorithm with a node-merging technique to address challenges in federated learning, offering solutions to improve efficiency, scalability, and robustness under diverse network and data conditions. And carefully calibrating the similarity threshold and the timing of merging is vital for optimizing the proposed method's performance in federated learning environments.

### Proposed algorithm for merging clients in FL

1: **Inputs:**
   local_models: A list of local model parameters from each node (N*M matrix where N = number of nodes, M = feature size for each model).
2: threshold: The correlation threshold above which nodes are considered similar (e.g., 0.7).
3: max_group_size: The maximum size of any group formed by merging nodes (e.g., 3).
4: **Output**:
5: groups: A list of groups, where each group contains the indices of merged nodes.
6: unmerged_nodes: A list of nodes that remain unchanged because they don't meet the merging criteria.
7: **Steps:**
8: **Calculate Pearson Correlation Matrix:**
9: For each pair of nodes (i, j), where i ≠ j:
10: Compute the Pearson correlation coefficient between local_models[i] and local_models[j].
11: $$PCC(x_i, x_j) = \frac{Cov(x_i, x_j)}{\sigma(x_i) \cdot \sigma(x_j)} \quad \forall i, j \in S, i \neq j$$
12: Store the result in correlation_matrix[i][j].
13: Result: A symmetric matrix where correlation_matrix[i][j] represents the similarity between node i and node j.
14: **Initialize:**
15: groups ← [] A list to hold groups of merged nodes.
16: used_nodes ← {} A set to keep track of nodes that have already been merged.
17: unmerged_nodes ← [] A list to hold nodes that remain independent.
18: **Group similar nodes:**
19: For each node i in correlation_matrix:
20: If i is in used_nodes, skip it (already processed).
21: Create a new group starting with i: group ← [i].
22: For each other node j (j ≠ i):
23: If j is not in used_nodes AND correlation_matrix[i][j] ≥ threshold:
24: Add j to group.
25: If the group size equals max_group_size, stop adding nodes.
26: If the group has more than one node:
27: Add the group to groups.
28: Mark all nodes in the group as used by adding them to used_nodes.
29: Otherwise, if the group has only one node (no matches), add i to unmerged_nodes.
30: **Handle remaining nodes:**
31: For each node i not in used_nodes:
32: Add i to unmerged_nodes (it remains unmerged).
33: **Return:**
34: groups: The list of merged groups.
35: unmerged_nodes: The list of nodes that were not merged with others.

### Scaffold algorithm using the proposed method for merging nodes

1: **Inputs:**
1: Initial global model: $x_0$
2: Global learning rate: $\eta_g$
3: Local learning rate: $\eta_l$
4: Number of communication rounds: $T$
5: Number of local epochs: $E$
6: Number of clients: $K$
7: Dataset size of client $i$: $n_i$
8: Total dataset size: $n = \sum_{i=1}^{K} n_i$
9: round in which the merging operation is performed
10: number of merge operations
11: number of nodes to merge
12: **Output:**
13: Final global model $x_T$
14: **Steps:**
15: Initialize global model $x_0$ and global control vector $c_0 = 0$.
16: For each client i:
17: Initialize local control vector $c_0^i = 0$
18: For t = 0 to T -1:  // Loop over communication rounds
19: **Server broadcasts:**
20: Global model $x_t$
21: Global control vector $c_t$
22: **Each client i performs:**
23: Receive $x_t$ and $c_t$ from the server.
24: Initialize local model: $x_t^i = x_t$
25: For e = 1 to E:  // Loop over local epochs
26: Compute local gradient:
27: $$\nabla f_i(x_t^i) = \frac{1}{n_i} \sum_{j \in \text{local data}} \nabla \ell(x_t^i; j),$$
28: where $\ell(x_t^i; j)$ is the loss for data point $j$. Update local model:
29: $$x_{t+1}^i = x_t^i - \eta_l \nabla f_i(x_t^i) + (c_t - c_t^i).$$
30: End For
31: Update local control vector:
34: $$c_{t+1}^i = c_t^i + (c_t - c_t^i) - \eta_l \nabla f_i(x_t^i).$$
35: Send $x_{t+1}^i$ and $c_{t+1}^i$ to the server.
36: **Server aggregates:**
37: Update global model:
38: $$x_{t+1} = x_t - \eta_g \sum_{i=1}^{K} \frac{n_i}{n}(x_{t+1}^i - x_t).$$
39: Update global control vector:
40: $$c_{t+1} = c_t + \frac{1}{K} \sum_{i=1}^{K} (c_{t+1}^i - c_t).$$
41: **Merging and create intermediary nodes:**

42:    If t = MERGE_ROUND do:
43:    run proposed merging algorithm
44:    Create intermediary nodes with the merged parameters.
45:    $x_{\text{merged}} = \alpha x_i + (1-\alpha)x_j, \quad \alpha \in [0,1]$
46:    $c_{\text{merged}} = \alpha c_i + (1-\alpha)c_j$
47:    Continue remaining rounds using the intermediary nodes and primary nodes.
48:    End For of Line 18.
49:    Return the final global model $x_T$.

In the proposed algorithm, the *used_nodes* list tracks nodes that have already been processed, ensuring no node is included in more than one group. The algorithm begins by selecting a single node i*i* as the initial member of a group and iteratively adds other nodes j*j* to the group if their correlation with i*i* exceeds a specified threshold and they have not been included in any other group. This approach ensures that nodes with high similarity are merged together into compact groups.

Nodes that fail to meet the similarity threshold with any other node or are not included in any group remain in the *unmerged_nodes* list as independent nodes, preserving their original state. This separation ensures that each node is either part of a meaningful group or left unaltered if no strong similarity exists.

The global model is updated on the server at each communication round which is obtained through the following formula

$$x_{t+1} = x_t - \eta_g \sum_{i=1}^{K} \frac{n_i}{n}(x_{t+1}^i - x_t) \quad (1)$$

Where $x_t$ is global model at communication round $t$, $x_{t+1}^i$ is Updated local model from client $i$, $\eta_g$ is global learning rate, $n_i$ is dataset size for client $i$ and $n = \sum_{i=1}^{K} n_i$ is total dataset size.

Each client updates its local model during local training as follows:

$$x_{t+1}^i = x_t^i - \eta_l \nabla f_i(x_t^i) + (c_t - c_t^i) \quad (2)$$

Where $x_t^i$ is local model for client $i$ at communication round $t$, $\nabla f_i(x_t^i)$ is gradient of the local loss function for client $i$, $c_t$ is global control vector, $c_t^i$ is local control vector for client $i$ and $\eta_l$ is local learning rate.

The server updates the global control vector by aggregating the local control by using following formula:

$$c_{t+1} = c_t + \frac{1}{K}\sum_{i=1}^{K}(c_{t+1}^i - c_t) \quad (3)$$

where $K$ is the number of participating clients.

In the next section, by testing different condition we will evaluate the robustness of proposed algorithm on the accuracy and convergence speed of the central model in federated learning scenario.

## V. EXPRIMNETAL RESULT

In order to evaluate the performance of the proposed algorithm and compare it with the scaffold algorithm in a federated learning scenario, we perform experiments. We assume that we have 10 local nodes and a central server. The experimental dataset is the MNIST database, which contains 60,000 training samples and 10,000 test samples. These samples are images of handwritten digits.

This dataset is used in most machine learning experiments. Here, to simulate the heterogeneity of the data, the samples are distributed among 10 local nodes in such a way that each node only includes a part of the classes of the dataset. In addition to the fact that the experiments are performed under normal network conditions, in order to examine the robustness of the algorithm, we apply two general network problems, including packet loss and network delay, in the testing process. We establish these conditions by not completing the training process in the epochs after the first epoch and by not fully training some local nodes.

The learning model that we use in the experiment is the CNN neural network, which is the most common architecture in deep learning applications and the round considered for merging local nodes is 4.

In federated learning applications, the main goal is to maintain data privacy in local nodes. On the other hand, each local node usually has only a part of the training examples related to the problem. Figure 1 shows the number of samples available in each of the 10 local nodes for our experiment. In order to simulate the Data heterogeneity conditions where each local node contains only a portion of the training samples belonging to some classes, we distribute the training samples of the MNIST dataset among the local nodes in such a way that each node contains only a portion of the examples belonging to each of the 10 classes. For example, for client 1, the number of samples belonging to each of the 0 to 10 classes is 5822, 622, 496, 6058, 0, 0, 261, 6086, 152 and 496 which is 19993 in total.

Figure 2 shows the comparison of the proposed method with the scaffold algorithm in different network conditions including normal mode, packet loss, and data poisoning attacks. As can be seen, after round 4, which is the round of merging local nodes based on the similarity of model parameters, the performance of the proposed method is better. This is because in situations where a number of models are targeted by a data poisoning attack and the overall learning process is being deviated, merging models first reduces the number of poisoned models and has less negative impact on the central model. This also applies to packet loss conditions. Because the number of models that have not been able to complete their training process due to the problem of packet loss is reduced through merging.

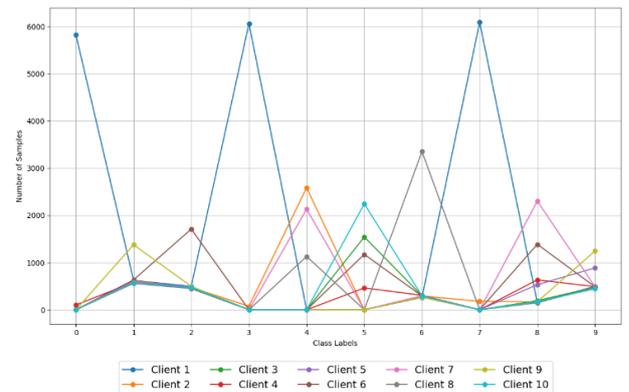

Fig. 1. Establishing Non-IID conditions for training samples from the MNIST dataset for 10 local clients

On the other hand, models that have not been attacked by a data poisoning attack or are not affected by the packet loss problem, since the data conditions are heterogeneous, the model parameters are not very similar to each other and fall below the threshold. As a result, after the merging step, they remain unmerged. This reduces the destructive role of the poisoned nodes and on the other hand, the healthy nodes continue to take over the overall learning process of the central model.

In other words In federated learning with heterogeneous data, the local models of nodes (even healthy nodes) have little similarity to each other due to differences in the training data. This difference is caused by things like: different data distributions (Non-IID), different data volumes per node, and diversity of features and labels.

In data heterogeneity, poisoned nodes are usually more similar to each other, provided that They have similar poisoned data and The attack target or poisoning strategy is the same.

If the attack conditions are more complex (e.g., poisoned nodes use different data), their models are less similar, but their destructive behavior to the central model still has a greater impact than healthy nodes.

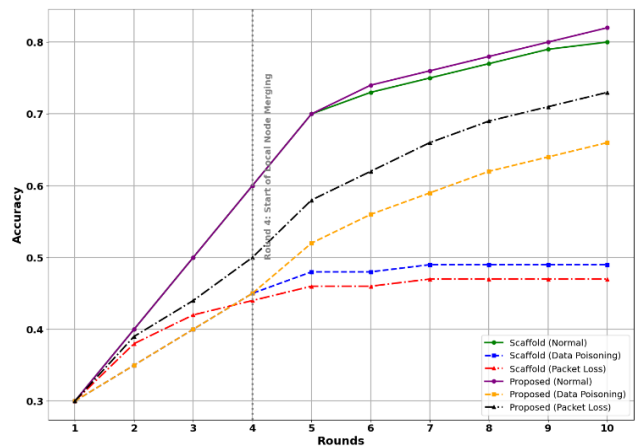

Fig. 2. Comparison of the proposed method with the scaffold algorithm in different network conditions

## VI. Conclusion

In this article, the challenges related to big data, including the 5V problem, are initially discussed. Subsequently, distributed machine learning is introduced as one of the approaches. A key aspect of distributed machine learning is federated learning, which emphasizes preserving the data privacy of local nodes. In this approach, data remains on the local device during the training process, and only the parameters of the learning model, such as weights and biases, are transmitted to the central model. However, federated learning algorithms face a challenge known as client drift, which can lead to low-quality local updates. This occurs when one or more clients deviate from the overall learning process and the central model due to factors like insufficient training data, data poisoning attacks, or packet loss, leading to a decrease in the accuracy of the central model.

The algorithm proposed in this article introduces a new concept called intermediary nodes. These nodes are created by merging two or more nodes with highly similar model parameters, thereby managing federated learning rounds in place of the original nodes. Nodes that do not exhibit such similarity remain unchanged. To assess the degree of similarity, we employ the correlation matrix and the Pearson correlation coefficient. When merging similar nodes, we use a threshold determined empirically.

The experimental results indicate that integrating the merging approach with the scaffold algorithm, which outperforms other federated learning algorithms under data heterogeneity conditions, yields relatively higher accuracy. This improvement is particularly notable when certain nodes experience data poisoning attacks or encounter packet loss issues.